\documentclass[conference]{IEEEtran}
\IEEEoverridecommandlockouts
\usepackage{amsmath,graphicx,algorithm,algorithmic,adjustbox,setspace}

\usepackage{xcolor}
\usepackage{enumitem}
\usepackage{hyperref}
\usepackage{multirow}
\usepackage{etoolbox,siunitx}
\robustify\bfseries
\usepackage{booktabs}
\usepackage{balance}
\usepackage{amssymb}
\usepackage{bm}
\usepackage{dsfont}
\usepackage{cite}

\newcommand{\red}[1]{\textcolor{red}{#1}}
\newcommand{\blue}[1]{\textcolor{blue}{#1}}

\let\OLDthebibliography\thebibliography
\renewcommand\thebibliography[1]{
  \OLDthebibliography{#1}
  \setlength{\parskip}{2pt}
  \setlength{\itemsep}{2pt plus 0.3ex}
}

\title{Robot Confirmation Generation and Action Planning Using Long-context Q-Former Integrated with Multimodal LLM}

\author{
    \IEEEauthorblockN{\textit{\shortstack{
        Chiori Hori$^1$, Yoshiki Masuyama$^1$, \\
        Siddarth Jain$^1$, Radu Corcodel$^1$, Devesh Jha$^1$, Diego Romeres$^1$, Jonathan Le Roux$^1$}
    }
    \vspace{.7\baselineskip}}        
    \IEEEauthorblockA{
        {$^{1}$Mitsubishi Electric Research Laboratories (MERL), Cambridge, MA, USA} \;
    }
}

\begin{document}
\maketitle
\begin{abstract}
Human-robot collaboration towards a shared goal requires robots to understand human action and interaction with the surrounding environment.
This paper focuses on human-robot interaction (HRI) based on human-robot dialogue that relies on the robot action confirmation and action step generation using multimodal scene understanding.
The state-of-the-art approach uses multimodal transformers to generate robot action steps aligned with robot action confirmation from a single clip showing a task composed of multiple micro steps.
Although actions towards a long-horizon task depend on each other throughout an entire video, the current approaches mainly focus on clip-level processing and do not leverage long-context information. This paper proposes a long-context Q-former incorporating left and right context dependency in full videos.
Furthermore, this paper proposes a text-conditioning approach
to feed text embeddings directly into the LLM decoder to mitigate the high abstraction of the information in text by Q-former.
Experiments with the YouCook2 corpus show that the accuracy of confirmation generation is a major factor in the performance of action planning.
Furthermore, we demonstrate that the long-context Q-former improves the confirmation and action planning by integrating VideoLLaMA3.
\end{abstract}

%

%%%%%%%%%%%%%%%%%%%%%%%%%%%%%%%%%%%%%%%%%%%%%%%%%%%%%%%%%%%%%%%%%%%%%%%%%%%%%%%%
\section{Introduction}
\label{sec:intro}
Human-robot interaction using natural language has the potential to be the most effective way for human-robot collaboration for shared tasks in daily life. Human-to-human collaboration is easily achieved because humans share the required knowledge about tasks and surrounding environments, and can understand other humans' behaviors. 
If there are unknown things, humans can use natural language to confirm what to do and how to do it.
Such fundamental functions for human-robot collaboration can be built using multimodal scene understanding that enables robots to interpret their environment and interact with humans based on such understanding.
This goal lies at the intersection of multiple avenues of research in speech understanding, audio event detection, object and action recognition in computer vision, physical sensing/manipulation for robot control, and natural language generation to interact with humans.
Achieving effective human-robot collaboration requires significant advances in the following areas: (1) multimodal scene understanding for human and robot environments, 
(2) robot action planning based on multimodal understanding and logical and physical affordance,  
(3) robot action generation/execution according to the planning, 
(4) robot action replanning if necessary, and (5) human-robot interaction via dialogue to achieve the goals efficiently.

There are several works on robotic manipulation actions at a high level that have proposed how instructions can be stored and analyzed~\cite{tenorth2013automated, yang2014cognitive, yang2015robot}.
Initial work utilized contrastive learning to learn a reward function to train reinforcement learning (RL) agents~\cite{sermanet2018time}. 
More recently, there have been some works 
using robot primitives and extraction of cost functions from task videos to enable imitation of demonstrated tasks~\cite{bahl2022human}.
There has also been some work on training perception modules on large sets of manipulation data to simplify learning of manipulation tasks~\cite{R3M_2022}. Finally, there has been growing interest in using vision and language for learning diverse robot skills. There are some works on training visual language models using human instruction videos that are well aligned to robot actions to generate action sequences \cite{Visualtranslating4robot2018,2D/3D_RN_Robot2021}.
Then, multimodal language models have been applied to robot action planning using various features, such as audio, visual, speech, and text, to expand knowledge acquisition of the tasks from human demonstration videos.
Initially, multimodal scene-understanding-based robot action planning applied an audio-visual transformer trained from cooking videos~\cite{hori23_interspeech}, where the action sequence was translated from human action descriptions.

\begin{figure*}[t]
    \centering
    \includegraphics[width=13.5cm]{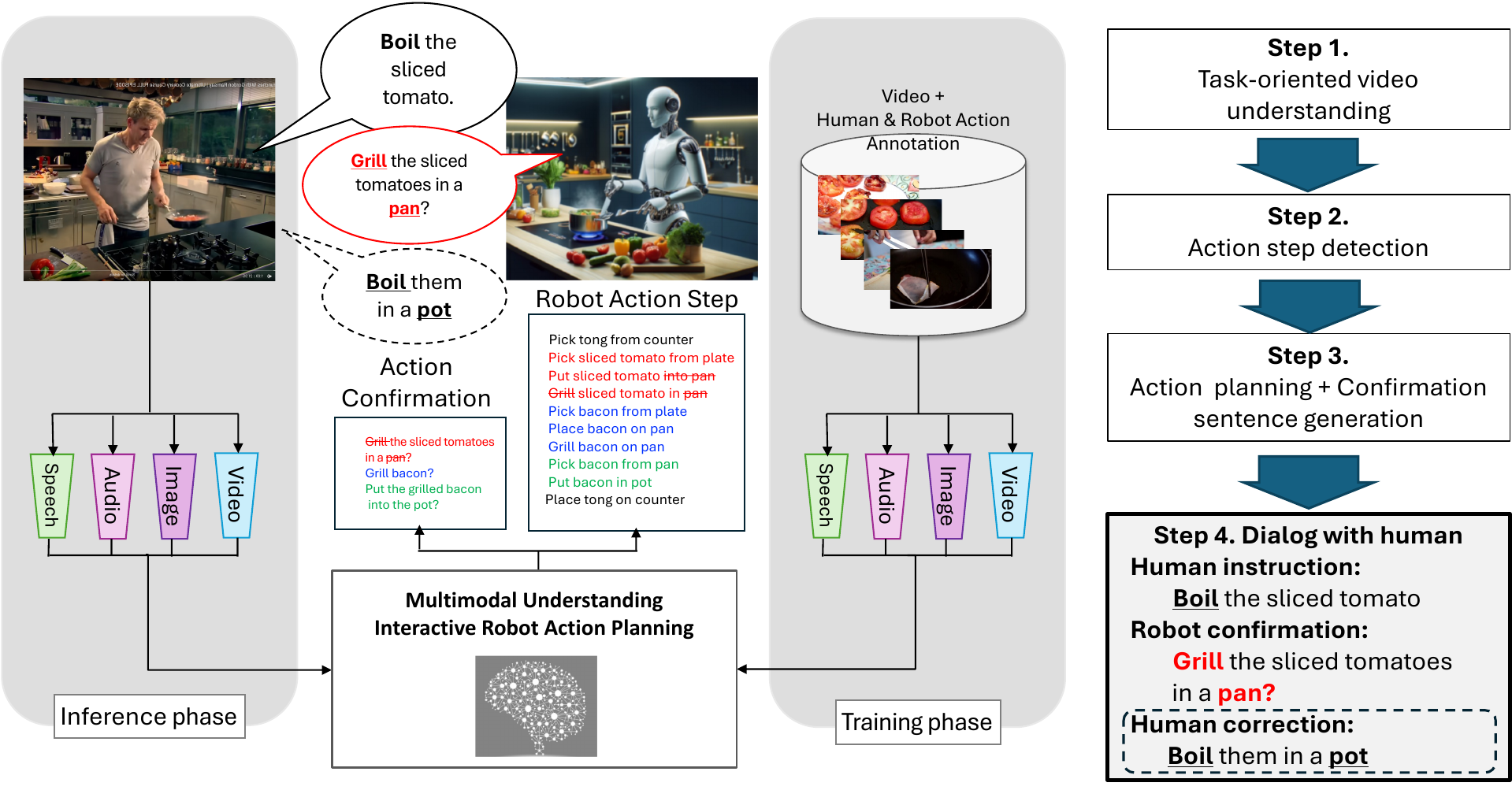}
    \vskip -3mm 
    \caption{Robot action confirmation sentences in natural language are simultaneously generated with micro-step action sequences. The multimodal features, such as videos, images, audio, and speech, are extracted from the human demonstration videos, and multimodal LLM models are trained to generate robot action descriptions to confirm whether the action is correct before taking the action. Although a humanoid robot is illustrated in this example, the action steps are designed for commoditized single-arm robots.}
    \label{fig:cookingconfirmation}
    \vskip -4mm
\end{figure*}

Recently, neural language models such as Large Language Models (LLMs) have been applied to bridge the gaps between sensing information and abstract-level understanding for robot action planning. 
This framework allows us to implement an end-to-end approach to build a human-robot collaboration system directly from multimodal scene understanding addressing all the steps (1) to (5). 
Promising use cases of LLMs were reported in the robotics research field, such as CLIPort \cite{cliport2021} and SayCan \cite{SayCan2022} in creating robotic agents that perform open-vocabulary tasks.
PROGPROMPT~\cite{singh2023progprompt} introduces a programmatic LLM prompt structure that facilitates the generation of plans in diverse environments, robot functionalities, and tasks. 
LLM-POP~\cite{Lingfeng_ICRA24} targets partially observable task planning,  where LLMs expand vocabulary and context considerations, while visual grounding LLMs enhance spatial reasoning capabilities. COWP~\cite{ding2023integrating} introduces an LLM-based open-world task planning system for robots.
Some works explore using LLMs to directly predict a dense sequence of end-effector poses for robot actions with vision models~\cite{kwon2024language}. 
Another study~\cite{raman2022planning} explores re-prompting strategies to enhance the executability and accuracy of LLM-generated plans but relies strictly on templated prompts.
Multimodal scene understanding has significantly advanced with AV-transformers, and has been applied to robot action planning with human action videos~\cite{hori23_interspeech}.
Despite its success, its capability for multimodal reasoning is still limited because it was impractical to cover all combinations of multimodal inputs from the existing videos.

To mitigate the discrepancy issues between the training and inference stages due to the data sparsity in multimodal fusion, BLIP2 \cite{BLIP2_ICML23}, narrowing down the semantic space, was applied to train a multimodal large language model (AVBLIP) for robot action planning \cite{Motonari_ICRA24_WS4Cooking}. 
Various features are embedded into the semantic space of LLMs by training a Q-former (cross-token transformer), 
which is an effective method in terms of computational efficiency and information retention compared to conventional fusion methods 
\cite{an2025llmcentricmultimodalfusionsurvey, wadekar2024evolutionmultimodalmodelarchitectures},
using both contrastive loss and action generation loss in this approach.
The Q-former-based multimodal LLM contributes to enhancing the performance of robot action planning.

However, automatic action planning errors are inevitable because of remaining discrepancies with the training environments and unexpected circumstances.
To avoid robots executing incorrect actions, a neural action replanning approach based on human error correction was proposed, where humans intervene to correct the action plans confirmed by robots using natural language before executing the planned actions~\cite{hori_2025_ICASSP}.
In this approach, the robot manipulation was segmented into a skill acquisition phase and a knowledge acquisition phase.
Although there have been some other research on language acquisition by robots to find associations between actions, objects, properties, and effects, and to map those associations to language~\cite{HRI_8616857, HRI_6082460}, it is unrealistic to train models handling a huge vocabulary using human demonstration videos through real human-robot interaction. 
To transfer (2) action planning trained from human action demonstration videos to (3) robot action generation/execution, a robot simulator was applied to generate and optimize executable robot actions~\cite{Kai_ICRA25_WS}.
In the next step, we build a system that can handle the real world beyond simulators~\cite{Stone_ICRA25_WS, zheng2025flarerobotlearningimplicit, li_2025_arxiv}.

In this research trend, this paper focuses on human-robot interaction (HRI) based on human-robot dialogue that relies on the robot action confirmation and action step generation.
The state-of-the-art approach leveraging AVBLIP generates robot action steps aligned with robot action confirmation from a single clip showing a task composed of multiple micro steps. Although action sequence dependency can be captured through an entire video to achieve the goal of a long-horizon task, the current single-clip-based approaches do not apply such long context information. This paper proposes a long-context Q-former incorporating left and right context dependency in full videos. Furthermore, the conventional AVBLIP feeds text embeddings into multimodal Q-former and could unexpectedly be degraded due to the high abstraction of the language information in the multimodal semantic space, thereby losing rich information represented in concrete words. To retain the original rich language information, this paper proposes a text-conditioning approach to feed text embeddings directly into the LLM decoder that generates robot action confirmation and robot action steps.
Finally, we show that the long-context Q-former improves the confirmation and action planning by integrating
VideoLLaMA3~\cite{damonlpsg2025videollama3}.

The main contributions of this work consist of (a) proposing a long-context Q-former for robot action confirmation sentence generation, 
(b) introducing a text conditioning that feeds text embedding vectors to an LLM decoder to retain original language information, (c) demonstrating the effectiveness of the proposed approaches for robot confirmation sentence and robot action sequence generation in the cooking domain, and
(d) examining the contribution of VideoLLaMA.
 
\section{Action Planning Data}
\noindent
\textit{a) Robot Action Description:}
YouCook2%
\footnote{http://youcook2.eecs.umich.edu/} 
is already annotated with human instructions in natural language to describe human cooking action steps, as introduced in \cite{zhou2018towards}. 
The cooking action steps for each video are annotated with start and end time stamps and English description. An example of the description is ``Grill the tomatoes in a pan and then put them on a plate" starting from 00:21 and ending at 00:52.

\noindent
\textit{b) Robot Micro-step Action:}
In the work of \cite{hori_2025_ICASSP},
the human instructions were translated into
a micro-step action sequence such that a single-arm robot could achieve the same goals as humans demonstrated actions, as illustrated in Fig.~\ref{fig:cookingconfirmation}. 
Although a humanoid robot is illustrated in this example, action steps are
designed for comoditized single-arm robots.
The micro-step action sequences for single-arm robot action are represented by ``single-arm action'', ``target object'', ``preposition'', and ``place'', to achieve the same actions as humans.
We borrowed the same data conditions used in \cite{hori_2025_ICASSP}.
Single-arm actions were selected from the following 12 candidates:
{Open, Close, Pick, Place, Pour, Stir, TurnOn, TurnOff, Wipe, Cut, Scoop, Squeeze}. The target objects were selected as one of the nouns in the human action instruction as much as possible.

\begin{figure}[t]
    \centering
    \includegraphics[width=9 cm]{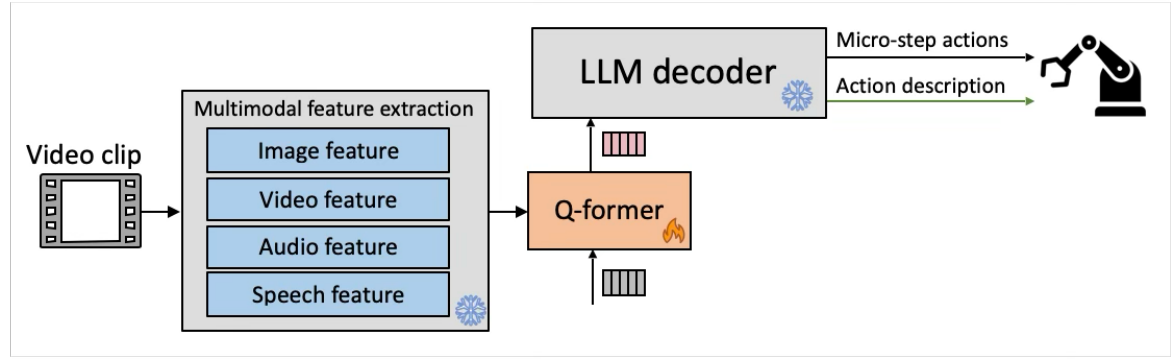}
    \vskip -3mm
     \caption{Q-former-based confirmation sentence generation and action planning model~\cite{hori_2025_ICASSP}. AVBLIP-based action generation with a Q-former. The generated embeddings are fed to the LLM decoder.}
    \label{fig:q-former}
    \vskip -4mm
\end{figure}

\begin{figure*}[t]
    \centering
    \includegraphics[width=13 cm]{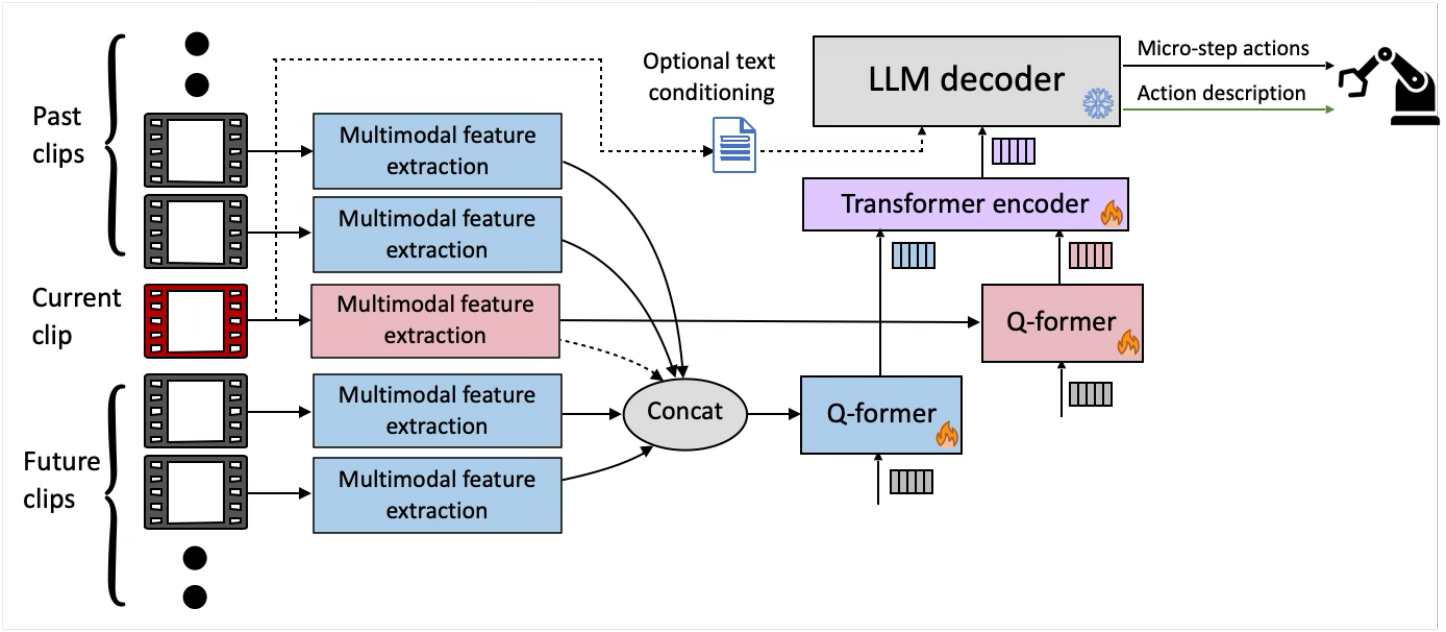}
     \vskip -3mm
     \caption{Long-context Q-former-based confirmation sentence generation and action planning model. AVBLIP-based action generation with two Q-former modules, one generates token embeddings from the current video clip and the other generates the embeddings from the surrounding video clips. The generated embeddings are combined with a transformer encoder and fed to the LLM decoder.}
    \label{fig:long-context-q-former}
    \vskip -4mm
\end{figure*}

\section{Robot action generation using AVBLIP}\label{sec:AVBLIP}
This work employs AVBLIP~\cite{Motonari_ICRA24_WS4Cooking}, where BLIP-2~\cite{BLIP2_ICML23}, a vision-language pre-training method, is extended to handle multimodal features. 
BLIP-2 bootstraps from a frozen image encoder and a frozen LLM, where a Querying Transformer (Q-former)~\cite{carion2020end} is trained to bridge the gap between the vision and text modalities. 
 The image encoder in BLIP-2 is replaced with audio-visual encoders for video, audio, and text feature sequences in AVBLIP.
Figure~\ref{fig:q-former} illustrates an architecture of AVBLIP consisting of a Q-former and an LLM decoder.
The Q-former is trained to extract a fixed number of embeddings from a multimodal encoder that outputs sequences with different lengths. 
The self-attention layers are shared between two transformer submodules: (1) a multimodal transformer that interacts with the frozen audio-visual encoders and (2) a text transformer that works as a text encoder and a text decoder. 
A set of learnable query embeddings is input to the multimodal transformer. 
The queries interact with each other through self-attention layers and interact with audio-visual features through cross-attention layers. 
The queries can additionally interact with the text through the same cross-attention layers. Finally, the queries are converted to an output feature.

The LLM Decoder generates a sequence of micro-step actions for a single-arm robot from multimodal features aligned to language features obtained by the Q-former. 
The LLM Decoder is constructed with a frozen LLM and a feed-forward layer. By using the LLM as a decoder, it leverages the LLM's inference capabilities when generating action sequences. In this study, we use OPT-2.7B~\cite{zhang2022opt} as the LLM.

The training of AVBLIP consists of two stages: (1) vision-language representation learning with frozen multimodal encoders and (2) vision-to-language generative learning with a frozen LLM.
In the second stage, we connect the Q-former to the frozen LLM Decoder and perform multimodal action sequence generation. As shown in Fig.~\ref{fig:q-former}, the extracted multimodal features from the video clip are converted to token embeddings using the Q-former. The embedding vectors are then projected to the LLM embedding space using a fully-connected layer. Then, the LLM Decoder generates action sequences and descriptions (confirmation sentences) for the given video clip. We use the cross-entropy loss function for the ground-truth sequences in this stage.

\section{Long-contex Q-former}
The AVBLIP model in Fig.~\ref{fig:q-former} can generate action sequences and confirmation sentences for short video clips of around 10-20 seconds. However, these clips are part of a longer instruction video, consisting of 5-10 sequential clips that aim at a single goal, e.g., ``cooking meatloaf''.
Therefore, incorporating contextual information from the previous and succeeding video clips is a promising extension of the model to generate more accurate sequences, because the video clips are interdependent: for example, some items cooked in a clip may be used in the next clip.

Figure~\ref{fig:long-context-q-former} shows an extended AVBLIP architecture based on a long-context Q-former, which contains two Q-former modules, one processes multi-modal features from the current clip as originally used and the other processes the features from the surrounding clips to utilize them to enhance the output for the current clip, where the surrounding clips could contain the current one.
Both Q-formers output token embeddings, which are then combined using a transformer encoder. 
Finally, the LLM decoder receives the combined embeddings to generate micro-step actions for the robot and action description (as a confirmation) for the human.
The two Q-formers and the transformer encoder are jointly trained in the same manner as the original AVBLIP framework described in Section \ref{sec:AVBLIP}.

\section{Text conditioning}
As demonstrated in \cite{Motonari_ICRA24_WS4Cooking,hori_2025_ICASSP}, the AVBLIP model can generate both micro-step action sequences and action descriptions by feeding query token embeddings to a pre-trained LLM, where the LLM will not further be trained or finetuned.
Thus, the Q-former is trained to generate token embeddings that convey the semantic information of the video clip as well as instructions to the LLM to generate a micro-step action sequence or an action description.
Although the set of embeddings is a compact and effective representation, it may not retain the detailed information necessary for the LLM to generate fine-grained action sequences.

In principle, directly feeding multi-modal features and a natural language instruction to the LLM could be more effective in generating accurate action sequences.
However, it is challenging to fine-tune the LLM to understand the multimodal features using limited training data and further to manually create prompts that generate specialized sequences, such as robot micro-step actions. In other words, the Q-former-based approach remains beneficial for obtaining suitable token embeddings, which serve as prompts for the pre-trained LLM.
However, there is a concern that precise semantic information could be lost. In particular, text information, such as speech subtitles obtained by automatic speech recognition, often contains exact keywords indicating specific ingredients and utensils. Therefore, it is reasonable that only the text information is fed to the LLM directly with minimal information loss.

This work applies text conditioning to the LLM by prepending the text information to the Q-former token embeddings as in Fig. \ref{fig:long-context-q-former}, where the text is tokenized and embedded by the same LLM.
The text can be subtitles in the video and/or automatic video descriptions generated by an external multimodal LLM such as VideoLLaMA \cite{damonlpsg2023videollama, damonlpsg2025videollama3, videoSALMONN}.

\section{Experiments}
\subsection{Setup}
Our proposed methods were tested using YouCook2 consisting of cooking action video clips aligned with human action instruction in natural language. Table \ref{tab:datasets} shows the data statistics. In this work, we used the validation set for evaluation since the test set was not publicly available.

\begin{table}[tb]
    \centering
    \caption{YouCookII~\cite{zhou2018towards}:
      ``Clip [sec]'' show the average clip duration.
    }
    \vspace{-2mm}\small 
    \label{tab:datasets}
    \begin{tabular}{lcccccc}
        \toprule
        & 
        \#Videos & 
        \#Clips& 
        Clip [sec] & 
         \\
        \midrule
        Training & 
        1173 &
        8743 &
        19.7 &
         \\
        Validation& 
        416 &
        3117 &
        19.8 &
         \\
        \bottomrule
    \end{tabular}
    \vskip -4mm
\end{table} 
We borrowed micro-step action sequences~\cite{hori_2025_ICASSP}.
The action set contains 2,790 unique phrases of 195 verbs, 2,229 objects, 33 prepositions, and 1002 places.

Multimodal features such as video, image, and audio were extracted using Omnivore \cite{girdhar2022omnivore}, Contrastive Language-Image Pre-Training (CLIP) \cite{radford2021learning}, and Audio Spectrogram Transformer (AST) \cite{gong21b_interspeech}, respectively. 
The image and video features were concatenated and projected to a single video feature sequence before feeding them to the encoder.
If a subtitle was available in the video, text features were extracted by Glove word embedding~\cite{pennington2014glove}. Otherwise, we fed an embedding vector for the \texttt{<unk>} label.
The numbers of dimensions of the audio, visual, and text features are $768$, $1024$, and $300$, respectively.

We initialized the Q-former with the pre-trained weights of BERTbase~\cite{devlin2018bert}, while the cross-attention layers were randomly initialized. We set the number of dimensions of the hidden layers to 768, which results in 188M parameters in total.
In the experiments, we used 32 queries, where each query has a 768-dimensional vector, which equals the hidden dimension of the Q-former.
For the long-context Q-former, we used 32 queries for each of the Q-former modules.
The transformer encoder to combine the contextual and current token embeddings had two transformer blocks, each of which had a self-attention layer and a full-connected layer with 768-dimensional hidden activations. These Q-formers and the transformer were jointly trained from scratch without any shared parameters.
For the text conditioning, speech subtitles annotated by YouTube and video descriptions obtained from VideoLLaMA3~\cite{damonlpsg2025videollama3} were fed into the LLM decoder.

The performance was evaluated using the BLEU-2 and METEOR scores computed between the generated and ground-truth sequences used in the robotics field \cite{Visualtranslating4robot2018,2D/3D_RN_Robot2021}.
We applied cross-validation, where one half of the validation set was used to select the best-epoch model, and the other half was used to measure model performance.
Thus, the shown scores are the average over the two subsets.

\subsection{Results}
\begin{table*}[t]
    \centering
    \small
    \caption{Examples of robot action confirmation sentence and robot action steps. \\ Object-level errors are highlighted in red. Corrected object names are highlighted in blue. If the corrected objects are supported by the subtitle, the VideoLLaMA3 description, and the context, they are also highlighted in blue.}
    \vskip -2mm
    \label{tab:example}
    \resizebox{.99\linewidth}{!}{
    \begin{tabular}{ll|l|l}
        \toprule
        Video clip Id    & Source    & Action sequence & Action description (confirmation) \\
        \midrule
        \multirow{6}{*}{sjh57ujp52M,6}
			    & Reference  & pick fish, place fish on towel  & remove the fish and drain the oil \\
                    & Baseline   & pick \red{spring roll}, place \red{spring roll} in oil & remove the \red{patties} from the oil and drain on paper towels \\
        	    & Proposed   & pick \blue{fish}, place \blue{fish} on paper \blue{towel} & remove the \blue{fish} from the pan and drain on the paper towels \\
                    \cmidrule{2-4}
			    & Subtitle   & \multicolumn{2}{l}{is take them off and I'm just going to place them on a little bit of paper tissue just to let them drain off so when you serve them up they're not all} \\
                    &             & \multicolumn{2}{l}{oily look at that} \\
                    \cmidrule{3-4}
                    & VideoLLaMA3 & \multicolumn{2}{l}{The video showcases the process of frying \blue{fish} in a pan. The \blue{fish} is initially placed into hot oil, where it sizzles and turns golden brown. Once cooked } \\
                    &             & \multicolumn{2}{l}{ to perfection, the \blue{fish} is lifted out with a spatula and transferred onto paper \blue{towels} to drain excess oil. This action takes place on a stove top, indicating} \\
                    &             & \multicolumn{2}{l}{that the cooking is taking place at home or in a small kitchen setting.} \\
        \midrule
        \multirow{6}{*}{Z5bpo2sBsl8,2}
        		& Reference  & pick knife from counter, cut cabbage with knife, place knife on counter  & continue chopping the cabbage \\
                    & Baseline   & pick \red{potato} from plate, place \red{potato} in bowl, pick bowl from counter, place bowl on counter, & grate some \red{parmesan cheese} on top of the \red{salad} \\
                    &            &  pick knife from counter, cut \red{potato} in bowl, place knife on counter  & \\
        	    & Proposed   & pick \blue{cabbage} from plate, place \blue{cabbage} on counter, pick knife from counter, cut \blue{cabbage} & chop the \blue{cabbage} \\
                    &            &  with knife, place knife on counter & \\
                    \cmidrule{2-4}
			    & Subtitle   & \multicolumn{2}{l}{What weight? Probably about 280. Wow. And I've had a... } \\
                    \cmidrule{3-4}
                    & VideoLLaMA3 & \multicolumn{2}{l}{In the video, a man and a woman are in a kitchen preparing food. The man is talking while standing next to the woman who is cutting \blue{cabbage} on a wooden board. } \\
                    \cmidrule{3-4}
                    & Context & \multicolumn{2}{l}{[previous] chop the \blue{cabbage} / [next] continue chopping the \blue{cabbage}} \\
        \bottomrule
    \end{tabular}
    }
    \vskip -5mm
\end{table*}

\begin{table}[t]
    \centering
    \caption{Quality of action sequences and descriptions generated by Long-context Q-former. The second column shows the context design. For example, ``-1,0,+1'' represents the context of previous (-1), current (0), and next (+1) clips. `*' denotes all previous or following clips.}
    \vskip -2mm
    \label{tab:result}
    \resizebox{\linewidth}{!}{
    \setlength{\tabcolsep}{6pt}
    \begin{tabular}{lccccc}
        \toprule
                       &  & \multicolumn{2}{c}{Action sequence} & \multicolumn{2}{c}{Action description} \\
        \cmidrule(lr){3-4}\cmidrule(lr){5-6}
        Model & Context &  BLEU-2 & METEOR & BLUE-2 & METEOR \\
        \midrule
        Baseline	 & -       & 0.370 & 0.260 & 0.229 & 0.159 \\
                     & -1,0	   & 0.374 & 0.264 & 0.235 & 0.166 \\
                     & -2,-1,0 &	0.376 & 0.265 & 0.237 & 0.168 \\
        Long-        & *,0	  & 0.376 & 0.266 & 0.237 & 0.169 \\
        Context      & -1,0,+1 & 0.379 & 0.269 & 0.240 & \bf 0.170 \\
                     & -2,-1,0,+1,+2 & \bf 0.381 & 0.269 & \bf 0.242 & \bf 0.170 \\
                     & *,0,*	  & \bf 0.381 & \bf 0.270 & 0.241 & 0.169 \\
        \midrule        
        Multitask\cite{hori_2025_ICASSP} & - & 0.370  & 0.257 & 0.220 & 0.158 \\
        ErrorCorrect\cite{hori_2025_ICASSP} & - & 0.375 & 0.258 & 0.231 & 0.161 \\
        \bottomrule    
    \end{tabular}
    }
    \vskip -4mm
\end{table}

Table~\ref{tab:result} shows the quality of action sequences and descriptions generated by the long-context Q-former.
``Baseline'' denotes the Q-former model trained with pairs of multimodal features and their target sequences without contextual information.
Unlike the method in \cite{hori_2025_ICASSP} that trains a single model to generate both action sequences and descriptions in a multitask manner, our baseline model was trained to generate action sequences and descriptions separately. Although we can use the same technique to achieve slightly better quality, we omitted that process for simplicity. In addition, our baseline model already achieves a quality comparable to the result of multitask training in \cite{hori_2025_ICASSP}. This could be because there are some differences in the hyperparameter setting, where we tuned the batch size, the learning rate, and the number of epochs in preliminary experiments.

We can see certain gains in BLEU and METEOR scores by using the long-context Q-former. The three rows after the baseline correspond to the results of left-context expansion, where introducing the previous two clips is sufficient, and we do not have to consider the full left context.
The next three rows display the results of introducing both left and right contexts, yielding further gains. Similar to the left-context expansion, it is sufficient to include two previous and two following clips to achieve the best quality.
These results demonstrate that the contextual information helps action and description generation and the long-context Q-former can effectively utilize that information.
Note that the training time of the long-context Q-former with full-video context (*,0,*) took 8.2 hours on a single A40 GPU, compared to 5.5 hours for the baseline, when performing two-stage training for 30 epochs with batch size 16. The average inference time was 5.8 seconds compared to 4.1 seconds for the baseline. Thus, the computational cost for the long-context model is not crucial.

\begin{table}[t]
    \centering
    \footnotesize
    \caption{Quality of action sequences and descriptions enhanced by text conditioning.
    }
    \vspace{-.2cm}
    \label{tab:result2}
    \resizebox{\linewidth}{!}{
    \setlength{\tabcolsep}{4pt}
    \begin{tabular}{lcccc}
        \toprule
             & \multicolumn{2}{c}{Action sequence} & \multicolumn{2}{c}{Action description} \\
        \cmidrule(lr){2-3}\cmidrule(lr){4-5}
        Conditioning    &  BLEU-2 & METEOR & BLEU-2 & METEOR \\
        \midrule
        Baseline (no conditioning) & 0.370	& 0.260 & 0.229 & 0.159 \\
        Speech subtitle	           & 0.411  & 0.281	& 0.257 & 0.170 \\
        VideoLLaMA3	               & 0.401  & 0.275 & 0.255 & 0.168 \\
        VideoLLaMA3+subtitle       & \bf 0.424 & \bf 0.290 & \bf 0.260 & \bf 0.178 \\
        Generated description	   & 0.398	& 0.272	& -	& - \\
        Ground-truth description   & 0.499	& 0.337	& -	& - \\
        \bottomrule    
    \end{tabular}
    }
    \vskip -4mm
\end{table}
Table \ref{tab:result2} shows the quality of action sequences and descriptions generated by the Q-former with text conditioning. 
First, speech subtitles provided a substantial gain from the baseline in all metrics.
This demonstrates that text conditioning of the LLM decoder is very effective for this task.
Then, we introduced a different text conditioning with video descriptions generated by the VideoLLaMA 7B model that achieves state-of-the-art performance in multiple video understanding benchmarks \cite{damonlpsg2025videollama3}.
To ask VideoLLaMA3 to generate rich video descriptions for text conditioning, we used a simple prompt ``Describe the cooking video in detail.''
This approach also improved the quality of the action sequences and descriptions, which are close to those with speech subtitles.
Then, we concatenated two text sequences and fed them to the LLM (``VideoLLaMA3+subtitle''), and obtained further improvement.
This result shows that rich and precise information is very useful when it is provided as an LLM's context.
Further, we tested the cases that use generated action descriptions and ground-truth descriptions.
With the descriptions generated by VideoLLaMA3+subtitle conditioning, the action sequence quality degraded. This is because the generated descriptions were much shorter than those of VideoLLaMA3+subtitle, and a certain level of information could be lost. 
The ground-truth descriptions were concise, yet they provided accurate descriptions, achieving the highest quality, which is considered the upper bound of the text conditioning approach.

\begin{table}[t]
    \centering
    \footnotesize
    \caption{Quality of action sequences and descriptions generated by long-context Q-former and text conditioning.
    }
    \vskip -2mm
    \label{tab:result3}
    \resizebox{\linewidth}{!}{
    \setlength{\tabcolsep}{4pt}
    \begin{tabular}{lccccc}
        \toprule
             & & \multicolumn{2}{c}{Action sequence} & \multicolumn{2}{c}{Action description} \\
        \cmidrule(lr){3-4}\cmidrule(lr){5-6}
        Context & Text conditioning    &  BLEU-2 & METEOR & BLEU-2 & METEOR \\
        \midrule
        Baseline	  &      -	                & 0.370 & 0.260 & 0.229 & 0.159 \\
        -  	          &   VideoLLaMA3+subtitle	& 0.424	& 0.290 & 0.260	& 0.178 \\
        -2,-1,0	      &   VideoLLaMA3+subtitle	& 0.427	& 0.290 & 0.264	& 0.180 \\
        -2,-1,0,+1,+2 &	  VideoLLaMA3+subtitle	& \bf 0.432	& \bf 0.291	& \bf 0.270	& \bf 0.183 \\

        \bottomrule    
    \end{tabular}
    }
    \vskip -2mm
\end{table}
Finally, we combined the long-context Q-former with text conditioning.
Table~\ref{tab:result3} shows the quality of the generated sequences using the combined approach. 
The results indicate that the performance gains by long-context Q-former and text conditioning are additive, achieving the best BLEU and METEOR scores.
The relative gains of the scores from the baseline are 16.7\% and 11.9\% for action sequences and 17.9\% and 15.1\% for action descriptions, respectively.

\section{Analysis and discussions}
Table~\ref{tab:example} shows examples of generated action sequences and descriptions, where the rows contain the reference, the baseline result, the sequence generated by the proposed method, the speech subtitle, and the VideoLLaMA3 description. They also include generated descriptions for the previous and the following clips in the second example. The proposed method used both long-context expansion and text conditioning.
In the first example, the errors in the baseline result are corrected by the proposed method, where the word ``spring roll'' is substituted with the correct word ``fish''. This shows that the VideoLLaMA description successfully supported the correct word through text conditioning.
In the second example, the object ``potato'' in the action sequence and ``parmesan cheese'' in the action description are corrected to ``cabbage'' using the proposed model. This correction could be made using information from the multimodal features of the previous and following clips, as the model predicted the word. Although the VideoLLaMA description also included ``cabbage'', the error could not be fixed when the long-context features were not used. We also have several examples in which speech subtitles support correcting the action sequences and descriptions, but we omit them due to space limitations.
As in those examples, the long context and text conditioning provide helpful information to generate accurate action sequences and descriptions.

\begin{table}[t]
    \centering
    \small
    \caption{Impact of text information on the feature side and the prompt side. ``Base'' denotes basic multimodal features we used in this work, i.e., image, video, audio, and subtitle features.}
    \label{tab:result4}
    \vskip -2mm
    \resizebox{\linewidth}{!}{
    \setlength{\tabcolsep}{4pt}
    \begin{tabular}{lccccc}
        \toprule
             & & \multicolumn{2}{c}{Action sequence} & \multicolumn{2}{c}{Action description} \\
        \cmidrule(lr){3-4}\cmidrule(lr){5-6}
        Features & Text conditioning    &  BLEU-2 & METEOR & BLEU-2 & METEOR \\
        \midrule
        Base               & -           & 0.370 & 0.260 & 0.229 & 0.159 \\
        Base - subtitle	   & -	         & 0.373 & 0.257 & 0.231 & 0.158 \\
        Base	           & Subtitle	 & 0.411 & 0.281 & 0.257 & 0.170 \\
        Base - subtitle	   & Subtitle	 & 0.409 & 0.281 & 0.258 & 0.170 \\
        Base + VideoLLaMA3 & -	         & 0.377 & 0.262 & 0.233 & 0.160 \\
        Base	           & VideoLLaMA3 & 0.401 & 0.275 & 0.255 & 0.168 \\
        Base + VideoLLaMA3 & VideoLLaMA3 & 0.400 & 0.276 & 0.256 & 0.170 \\
        Subtitle only & Subtitle & 0.223 & 0.219 & 0.174 & 0.126 \\
        VideoLLaMA3	only & VideoLLaMA3 & 0.353 & 0.252 & 0.211 & 0.151 \\
        \bottomrule    
    \end{tabular}
    }
    \vskip -3mm
\end{table}
Finally, we show the impact of text information on the feature side and the prompt side, i.e., text conditioning.
Table \ref{tab:result4} shows the quality of action sequences and descriptions in different feature combinations.
The first row shows the baseline result, where we used ``base'' features consisting of image, video, audio, and subtitle features.
The second row shows the result when we removed subtitles from the base features. Compared to the baseline, the BLEU and METEOR scores do not change significantly, meaning that the baseline model did not effectively utilize the subtitle feature.
The third row indicates that text conditioning with subtitles substantially improves the sequence quality. 
The fourth row corresponds to the result when we use subtitles on both sides, showing that adding subtitles to the features does not help.
We also used VideoLLaMA descriptions as features and/or text conditioning, and obtained similar results to the case of subtitles.
The last two rows show the results when we removed the base features, where we used only subtitles or VideoLLaMA descriptions, because the text may already contain rich semantic information of the video, and therefore, we want to check if the base features are still necessary for this task.
We can see substantial degradation of the quality, which means that the multimodal features are still important for this task even though a subtitle or rich video description is provided.
Furthermore, we found that not all videos have valid subtitles. They include transcripts of unrelated utterances, non-English subtitles, and background music alone, which may not contribute to the performance. Therefore, the multimodal features are essential to compensate for invalid subtitles.

\section{Conclusions}
This paper proposed a method for robot action sequence and confirmation sentence generation that leverages (a) a long-context Q-former considering left and right context dependency in full videos and (b) a text-conditioned LLM decoder to retain the precise language information.
We trained the above proposed models, generating single-arm robot micro-step action sequences and robot action confirmation in natural language using the YouCook2 dataset.
To mitigate the sparseness of the human action descriptions, we leveraged the video descriptions obtained from VideoLLaMA3. 
Experimental results show that our proposed long-context Q-former outperformed the baseline model for all metrics. 
We confirmed that the combination of all multimodal features and text-conditioning performed the best.
Future work includes (1) multitask training of robot action steps and action descriptions, (2) comparison with other multimodal fusion approaches such as simple sequence concatenation or pooling+MLP, (3) evaluation for tasks other than cooking, and (4) evaluation using a simulator or a real robot.

\clearpage
\balance
\bibliographystyle{IEEEtran}
\bibliography{mybib}

@string{icassp = "Proc. ICASSP"}

@string{interspeech = "Proc. Interspeech"}

@string{icml = "Proc. ICML"}

@string{cvpr = "Proc. CVPR"}

@string{aaai = "Proc. AAAI"}

@string{emnlp = "Proc. EMNLP"}

@string{eccv = "Proc. ECCV"}

@string{neurips = "Proc. NeurIPS"}

@string{icra = "Proc. ICRA"}

@string{iros = "Proc. IROS"}

@misc{wadekar2024evolutionmultimodalmodelarchitectures,
      title={The Evolution of Multimodal Model Architectures}, 
      author={Shakti N. Wadekar and Abhishek Chaurasia and Aman Chadha and Eugenio Culurciello},
      year={2024},
      eprint={2405.17927},
      archivePrefix={arXiv},
      primaryClass={cs.AI},
      url={https://arxiv.org/abs/2405.17927}, 
}

@misc{an2025llmcentricmultimodalfusionsurvey,
      title={Towards LLM-Centric Multimodal Fusion: A Survey on Integration Strategies and Techniques}, 
      author={Jisu An and Junseok Lee and Jeoungeun Lee and Yongseok Son},
      year={2025},
      eprint={2506.04788},
      archivePrefix={arXiv},
      primaryClass={cs.CL},
      url={https://arxiv.org/abs/2506.04788}, 
}

@inproceedings{Kai_ICRA25_WS,
  author={Lu, Kai and Ma, Chenyang and Hori, Chiori and Romeres, Diego},
  title={KitchenVLA: Iterative Vision-Language Corrections for Robotic Execution of Human Tasks},
  year=2025,
  booktitle={Proc. ICRA Workshop for ``Safely Leveraging Vision-Language Foundation Models in Robotics :
Challenges and Opportunities''},
  pages={},
  doi={}
}

@article{zheng2025flarerobotlearningimplicit,
      title={FLARE: Robot Learning with Implicit World Modeling}, 
      author={Ruijie Zheng and Jing Wang and Scott Reed and Johan Bjorck and Yu Fang and Fengyuan Hu and Joel Jang and Kaushil Kundalia and Zongyu Lin and Loic Magne and Avnish Narayan and You Liang Tan and Guanzhi Wang and Qi Wang and Jiannan Xiang and Yinzhen Xu and Seonghyeon Ye and Jan Kautz and Furong Huang and Yuke Zhu and Linxi Fan},
      year={2025},
      journal={arXiv preprint arXiv:2505.15659},
}

@inproceedings{Stone_ICRA25_WS,
  author={Tao, Stone and others},
  title={MANISKILL3: {GPU} PARALLELIZED SIMULATION AND RENDERING FOR GENERALIZABLE EMBODIED {AI}},
  year=2025,
  booktitle={Proc. ICRA Workshop for ``7th Robot Learning Workshop: Towards Robots with Human-Level Abilities''},
  pages={},
  doi={}
}

@article{li_2025_arxiv,
      title={{WorldEval}: World Model as Real-World Robot Policies Evaluator}, 
      author={Yaxuan Li and Yichen Zhu and Junjie Wen and Chaomin Shen and Yi Xu},
      year={2025},
      journal={arXiv preprint arXiv:2505.19017},
}

@ARTICLE{HRI_8616857,
  author={Saponaro, Giovanni and Jamone, Lorenzo and Bernardino, Alexandre and Salvi, Giampiero},
  journal={IEEE Transactions on Cognitive and Developmental Systems}, 
  title={Beyond the Self: Using Grounded Affordances to Interpret and Describe Others’ Actions}, 
  year={2020},
  volume={12},
  number={2},
  pages={209-221},
  doi={10.1109/TCDS.2018.2882140}}

@ARTICLE{HRI_6082460,
  author={Salvi, Giampiero and Montesano, Luis and Bernardino, Alexandre and Santos-Victor, José},
  journal={IEEE Transactions on Systems, Man, and Cybernetics, Part B (Cybernetics)}, 
  title={Language Bootstrapping: Learning Word Meanings From Perception–Action Association}, 
  year={2012},
  volume={42},
  number={3},
  pages={660-671},
  doi={10.1109/TSMCB.2011.2172420}}

@inproceedings{carion2020end,
  title={End-to-end object detection with transformers},
  author={Carion, Nicolas and Massa, Francisco and Synnaeve, Gabriel and Usunier, Nicolas and Kirillov, Alexander and Zagoruyko, Sergey},
  booktitle=eccv,
  year={2020},
}

@inproceedings{hori23_interspeech,
  author={Chiori Hori and Puyuan Peng and David Harwath and Xinyu Liu and Kei Ota and Siddarth Jain and Radu Corcodel and Devesh Jha and Diego Romeres and Jonathan {Le Roux}},
  title={{Style-transfer based Speech and Audio-visual Scene understanding for Robot Action Sequence Acquisition from Videos}},
  year=2023,
  booktitle=interspeech,
  pages={4663--4667},
  doi={10.21437/Interspeech.2023-1983}
}

@inproceedings{Motonari_ICRA24_WS4Cooking,
  author={Motonari Kambara and Komei Sugiura and Sammer Khurana and Kei Ota and Siddarth Jain and Radu Corcodel and Devesh Jha and Diego Romeres and Jonathan {Le Roux} and Chiori Hori
  },
  title={Human Action understanding-based Robot Planning using Multimodal {LLM}},
  year=2024,
  booktitle={Proc. ICRA Workshop for ``Cooking Robotics: Perception and motion planning''},
  pages={},
  doi={}
}

@inproceedings{Lingfeng_ICRA24,
  author={Sun, Lingfeng and Jha, Devesh K and Hori, Chiori and Jain, Siddarth and Corcodel, Radu and Zhu, Xinghao and Tomizuka, Masayoshi and Romeres, Diego},
  title={Interactive Planning Using Large Language Models for Partially Observable Robotic Tasks},
  year=2024,
  booktitle=icra,
  pages={},
  doi={}
}

@inproceedings{hori_2025_ICASSP,
  title={Interactive robot action replanning using multimodal llm trained from human demonstration videos},
  author={Hori, Chiori and Kambara, Motonari and Sugiura, Komei and Ota, Kei and Khurana, Sameer and Jain, Siddarth and Corcodel, Radu and Jha, Devesh and Romeres, Diego and Le Roux, Jonathan},
  booktitle=icassp,
  year={2025},
}

@article{ding2023integrating,
  title={Integrating action knowledge and {LLMs} for task planning and situation handling in open worlds},
  author={Ding, Yan and Zhang, Xiaohan and Amiri, Saeid and Cao, Nieqing and Yang, Hao and Kaminski, Andy and Esselink, Chad and Zhang, Shiqi},
  journal={Autonomous Robots},
  volume={47},
  number={8},
  pages={981--997},
  year={2023},
  publisher={Springer}
}

@article{kwon2024language,
  title={Language models as zero-shot trajectory generators},
  author={Kwon, Teyun and Di Palo, Norman and Johns, Edward},
  journal={IEEE Robotics and Automation Letters},
  year={2024},
}

@inproceedings{BLIP2_ICML23,
author = {Li, Junnan and Li, Dongxu and Savarese, Silvio and Hoi, Steven},
title = {{BLIP-2}: Bootstrapping language-image pre-training with frozen image encoders and large language models},
year = {2023},
booktitle = icml,
}

@inproceedings{bahl2022human,
    title={Human-to-Robot Imitation in the Wild},
    author={Bahl, Shikhar and Gupta, Abhinav and Pathak, Deepak},
    booktitle={Proc. RSS},
    year={2022}
  }

@inproceedings{singh2023progprompt,
  title={Progprompt: Generating situated robot task plans using large language models},
  author={Singh, Ishika and Blukis, Valts and Mousavian, Arsalan and Goyal, Ankit and Xu, Danfei and Tremblay, Jonathan and Fox, Dieter and Thomason, Jesse and Garg, Animesh},
  booktitle=icra,
  year={2023},
}

@inproceedings{raman2022planning,
  title={Planning with large language models via corrective re-prompting},
  author={Raman, Shreyas Sundara and Cohen, Vanya and Rosen, Eric and Idrees, Ifrah and Paulius, David and Tellex, Stefanie},
  booktitle={NeurIPS Foundation Models for Decision Making Workshop},
  year={2022}
}

@article{R3M_2022,
  author = {Nair, Suraj and Rajeswaran, Aravind and Kumar, Vikash and Finn, Chelsea and Gupta, Abhinav},
  title = {{R3M}: A Universal Visual Representation for Robot Manipulation},
  journal = {arXiv preprint arXiv:2203.12601},
  year = {2022},
}

@INPROCEEDINGS{2D/3D_RN_Robot2021,
  author={Xu, Xin and Qian, Kun and Zhou, Bo and Chen, Shenghao and Li, Yitong},
  booktitle=icra, 
  title={Two-stream {2D/3D} Residual Networks for Learning Robot Manipulations from Human Demonstration Videos}, 
  year={2021},}

@INPROCEEDINGS{Visualtranslating4robot2018,
  author={Nguyen, Anh and Kanoulas, Dimitrios and Muratore, Luca and Caldwell, Darwin G. and Tsagarakis, Nikos G.},
  booktitle=icra, 
  title={Translating Videos to Commands for Robotic Manipulation with Deep Recurrent Neural Networks}, 
  year={2018},}

@inproceedings{cliport2021,
  title     = {{CLIPort}: What and Where Pathways for Robotic Manipulation},
  author    = {Shridhar, Mohit and Manuelli, Lucas and Fox, Dieter},
  booktitle = {Proc. CoRL},
  year      = {2021},
}

@inproceedings{yang2015robot,
  title={Robot learning manipulation action plans by ``watching" unconstrained videos from the world wide web},
  author={Yang, Yezhou and Li, Yi and Fermuller, Cornelia and Aloimonos, Yiannis},
  booktitle=aaai,
  year={2015}
}

@inproceedings{tenorth2013automated,
  title={Automated alignment of specifications of everyday manipulation tasks},
  author={Tenorth, Moritz and Ziegltrum, Johannes and Beetz, Michael},
  booktitle=iros,
  pages={5923--5928},
  year={2013},
}

@article{yang2014cognitive,
  title={A cognitive system for understanding human manipulation actions},
  author={Yang, Yezhou and Guha, Anupam and Fermuller, C and Aloimonos, Yiannis},
  journal={Advances in Cognitive Systems},
  volume={3},
  pages={67--86},
  year={2014}
}

@article{SayCan2022,
  author = {Ahn, Michae and Brohan, Anthony and Brown, Noah and Chebotar, Yevgen and Cortes, Omar and David, Byron and Finn, Chelsea and Fu, Chuyuan and Gopalakrishnan, Keerthana and Hausman, Karol and Herzog, Alex and Ho, Daniel and Hsu, Jasmine and Ibarz, Julian and Ichter, Brian and Irpan, Alex and Jang, Eric and Ruano, Rosario Jauregui and Jeffrey, Kyle and Jesmonth, Sally and Joshi, Nikhil J and Julian, Ryan and Kalashnikov, Dmitry and Kuang, Yuheng and Lee, Kuang-Huei and Levine, Sergey and Lu, Yao and Luu, Linda and Parada, Carolina and Pastor, Peter and Quiambao, Jornell and Rao, Kanishka and Rettinghouse, Jarek and Reyes, Diego and Sermanet, Pierre and Sievers, Nicolas and Tan, Clayton and Toshev, Alexander and Vanhoucke, Vincent and Xia, Fei and Xiao, Ted and Xu, Peng and Xu, Sichun and Yan, Mengyuan and Zeng, Andy },
  title = {Do As {I} Can, Not As {I} Say: Grounding Language in Robotic Affordances},
  journal = {arXiv preprint arXiv:2204.01691},
  year = {2022},
}

@inproceedings{girdhar2022omnivore,
  title={{Omnivore: A Single Model for Many Visual Modalities}},
  author={Girdhar, Rohit and Singh, Mannat and Ravi, Nikhila and van der Maaten, Laurens and Joulin, Armand and Misra, Ishan},
  booktitle=cvpr,
  year={2022}
}

@inproceedings{radford2021learning,
  title={Learning transferable visual models from natural language supervision},
  author={Radford, Alec and Kim, Jong Wook and Hallacy, Chris and Ramesh, Aditya and Goh, Gabriel and Agarwal, Sandhini and Sastry, Girish and Askell, Amanda and Mishkin, Pamela and Clark, Jack and others},
  booktitle=icml,
  pages={8748--8763},
  year={2021},
}

@inproceedings{gong21b_interspeech,
  author={Yuan Gong and Yu-An Chung and James Glass},
  title={{AST: Audio Spectrogram Transformer}},
  year=2021,
  booktitle=interspeech,
  pages={571--575},
  doi={10.21437/Interspeech.2021-698}
}

@article{zhang2022opt,
  title={{OPT: Open Pre-trained Transformer Language Models}},
  author={Zhang, Susan and Roller, Stephen and Goyal, Naman and Artetxe, Mikel and Chen, Moya and Chen, Shuohui and Dewan, Christopher and others},
  journal={arXiv preprint arXiv:2205.01068},
  year={2022}
}

@inproceedings{pennington2014glove,
  title={Glove: Global vectors for word representation},
  author={Pennington, Jeffrey and Socher, Richard and Manning, Christopher D},
  booktitle=emnlp,
  pages={1532--1543},
  year={2014}
}

@inproceedings{zhou2018towards,
  title={Towards automatic learning of procedures from web instructional videos},
  author={Zhou, Luowei and Xu, Chenliang and Corso, Jason J},
  booktitle=aaai,
  year={2018}
}

@inproceedings{sermanet2018time,
  title={Time-contrastive networks: Self-supervised learning from video},
  author={Sermanet, Pierre and Lynch, Corey and Chebotar, Yevgen and Hsu, Jasmine and Jang, Eric and Schaal, Stefan and Levine, Sergey and Brain, Google},
  booktitle=icra,
  year={2018},
}

@article{devlin2018bert,
  title={{BERT}: Pre-training of deep bidirectional transformers for language understanding},
  author={Devlin, Jacob and Chang, Ming-Wei and Lee, Kenton and Toutanova, Kristina},
  journal={arXiv preprint arXiv:1810.04805},
  year={2018}
}

@article{damonlpsg2023videollama,
  title = {Video-{LLaMA}: An Instruction-tuned Audio-Visual Language Model for Video Understanding},
  author = {Zhang, Hang and Li, Xin and Bing, Lidong},
  journal = {arXiv preprint arXiv:2306.02858},
  year = {2023},
}

@article{damonlpsg2025videollama3,
  title={{VideoLLaMA} 3: Frontier Multimodal Foundation Models for Image and Video Understanding},
  author={Zhang, Boqiang and Li, Kehan and Cheng, Zesen and Hu, Zhiqiang and Yuan, Yuqian and Chen, Guanzheng and Leng, Sicong and Jiang, Yuming and Zhang, Hang and Li, Xin and Jin, Peng and Zhang, Wenqi and Wang, Fan and Bing, Lidong and Zhao, Deli},
  journal={arXiv preprint arXiv:2501.13106},
  year={2025},
}

@inproceedings{videoSALMONN,
author = {Sun, Guangzhi and Yu, Wenyi and Tang, Changli and Chen, Xianzhao and Tan, Tian and Li, Wei and Lu, Lu and Ma, Zejun and Wang, Yuxuan and Zhang, Chao},
title = {video-{SALMONN}: {Speech}-enhanced audio-visual large language models},
year = {2024},
booktitle = "Proc. ICML",
}
\end{document}